\documentclass{article}

     \usepackage[final]{icbinb_neurips_2020}

\usepackage[utf8]{inputenc} 
\usepackage[T1]{fontenc}    
\usepackage{hyperref}       
\usepackage{url}            
\usepackage{booktabs}       
\usepackage{amsfonts}       
\usepackage{nicefrac}       
\usepackage{microtype}      

\usepackage{amsmath}
\usepackage{dsfont}
\usepackage{graphicx}
\usepackage{wrapfig}

\title{Uses and Abuses of the Cross-Entropy Loss: \\ Case Studies in Modern Deep Learning}

\author{%
  Elliott Gordon-Rodriguez 
  \\
  Department of Statistics\\
  Columbia University\\
  \texttt{eg2912@columbia.edu} \\
  \And
  Gabriel Loaiza-Ganem \\
  Layer6 AI \\
  \texttt{gabriel@layer6.ai} \\
  \AND
  Geoff Pleiss \\
  Zuckerman Institute\\
  Columbia University\\
  \texttt{gmp2162@columbia.edu} \\
  \And
  John P. Cunningham \\
  Department of Statistics\\
  Columbia University\\
  \texttt{jpc2181@columbia.edu} \\
}

\begin{document}

\maketitle

\begin{abstract}
Modern deep learning is primarily an experimental science, in which empirical advances occasionally come at the expense of probabilistic rigor.
Here we focus on one such example; namely the use of the categorical cross-entropy loss to model data that is not strictly categorical, but rather takes values on the simplex.
This practice is standard in neural network architectures with label smoothing and actor-mimic reinforcement learning, amongst others.
Drawing on the recently discovered {continuous-categorical} distribution, we propose probabilistically-inspired alternatives to these models,
providing an approach that is more principled and theoretically appealing.
Through careful experimentation, including an ablation study, we identify the potential for outperformance in these models, thereby highlighting the importance of a proper probabilistic treatment, as well as illustrating some of the failure modes thereof.\footnote{Our code is available at \url{https://github.com/cunningham-lab/cb_and_cc}.}
\end{abstract}

\section{Introduction}

The cross-entropy loss is one of the most commonly used loss functions for training deep neural network models, most notably in (multi-class) classification problems.
When applied to categorical data, this loss function corresponds to a probabilistic log-likelihood, therefore resulting in favorable estimation properties.
On the other hand, several prominent methods in modern machine learning are concerned with fitting data that is not quite categorical, but simplex-valued; key examples being the ``soft targets'' in label smoothing (LS) \cite{szegedy2016rethinking}, and ``expert policies'' in actor-mimic reinforcement learning (AMN) \cite{parisotto2015actor}, amongst others \cite{hinton2015distilling, tzeng2015simultaneous}.
In these methods, the deep learning community has defaulted to borrowing the same cross-entropy loss from the categorical case, despite the fact that it no longer defines a bona fide probability model.
As well as highlighting this practice and putting it into question, our work proposes adjusting the LS and AMN objective functions by replacing the cross-entropy loss with the log-likelihood of the recently discovered \emph{continuous-categorical} (CC) distribution \cite{gordon2020continuous}.
Doing so amounts to incorporating a normalizing constant to our model, or in other words, adding the factor that scales the cross-entropy loss to a valid probability density function over the simplex.
As of yet, such an approach has only been considered in the context of knowledge distillation \cite{gordon2020continuous}, although the one-dimensional special case (corresponding to the binary cross-entropy with $[0,1]$-valued data) has been studied more extensively \cite{loaiza2019continuous}.
Our inspiration draws from both of these works, although our focus is primarily on LS and AMN architectures instead.

Our exposition is organized as follows (note that our two main sections, \ref{sec:ls} and \ref{sec:rl}, are based on the same idea, but are broadly independent of one another and can be read separately):
\begin{itemize}
  \item In section \ref{sec:background} we detail the relevant background on the continuous-categorical distribution, highlighting its close connection to the cross-entropy loss. 
  \item Section \ref{sec:ls} focuses on label smoothing. We propose a novel CC-LS model and perform an ablation study to isolate its potential as a regularizer for classification networks, as well as a qualitative assessment of its learned representations.
  \item Section \ref{sec:rl} focuses on actor-mimic reinforcement learning. We recast the AMN model as the solution to a regression problem of simplex-valued data, we propose a novel CC-AMN model, and we provide an experimental evaluation thereof.
  \item Section \ref{sec:conclusion} concludes, combining insights from CC-LS and CC-AMN, and discussing potential directions for future research.
\end{itemize}

\section{Background} \label{sec:background}

We preface the introduction of the continuous-categorical distribution with a brief notational overview of the categorical cross-entropy loss, which will highlight the close connection between the two and will provide an orthogonal viewpoint to its original presentation in \cite{gordon2020continuous}.


Categorical data refers to observations $y$ that take values in a discrete sample space $\Omega$ formed by $K$ distinct elements, which are typically expressed using the $K$ one-hot vectors that form the standard basis of $\mathbb R^K$, namely $\Omega =  \{ e_1, \dots e_K \}$, where $(e_k)_j = \mathds{1} ( k = j )$.
In this notation, the cross-entropy loss is equivalent to the negative log-likelihood of $y \in \Omega$ under a categorical distribution with parameter $\pi$: 
\begin{align} \label{eq:xe_loss}
 l ( \pi ; y ) = - \sum_{k=1}^K y_k \log \pi_k \iff p(y;\pi) = \prod_{k=1}^K \pi_k^{y_k}.
\end{align}
In other words, the cross-entropy loss defines a coherent probabilistic model for discrete data over $K$ classes.
This elementary fact should not be overlooked; it provides the benefits of the theory of maximum likelihood estimation, including frequentist consistency and asymptotic efficiency, as well as enabling efficient Bayesian inference by specifying a conjugate prior.

\subsection{From the Cross-Entropy to the Continuous-Categorical} 

So far so good. 
However, what happens when the observation is not quite categorical, but instead takes values on the simplex, $\Delta^K = \{ y \in \mathbb R_+^K : \sum_{k=1}^K y_k = 1 \}$? 
Such data is called \emph{compositional}, and is common in the sciences \cite{aitchison1982statistical}. 
In the deep learning literature, while not explicitly referred to as such, compositional data plays a key role in label smoothing  \cite{szegedy2016rethinking}, actor-mimic reinforcement learning \cite{parisotto2015actor}, {knowledge distillation} \cite{hinton2015distilling}, and domain adaptation \cite{tzeng2015simultaneous}.
In all of these methods, neural networks are trained to target a simplex-valued outcome, $y \in \Delta^K$, using the cross-entropy loss $l(\lambda ; y) = - \sum_{k=1}^K y_k \log \lambda_k$, where $\lambda$ represents the output of a neural network. 
Crucially though, the change in sample space from $\Omega$ to $\Delta^K$ breaks the equivalence in (\ref{eq:xe_loss}) because the right-hand expression no longer defines a proper probability distribution; its integral over $\Delta^K$ does not normalize to 1. 

Given the attractive properties of maximum likelihood estimation, there are still good reasons why a legitimate probability model is desirable (see \cite{loaiza2019continuous} for a more detailed discussion).
The classical statistics literature offers some possibilities, notably the use of \emph{logratios} \cite{aitchison1982statistical,  aitchison1994principles, aitchison1999logratios, egozcue2003isometric}, or \emph{Dirichlet regression} \cite{campbell1987multivariate, hijazi2009modelling}.
However, we argue that the most natural probabilistic solution is to apply the recently discovered {continuous-categorical} distribution \cite{gordon2020continuous}, since this corresponds to normalizing the cross-entropy loss directly so that it becomes a genuine log-likelihood model, 
namely:
\begin{align} \label{eq:cts_xe_loss}
 l ( \lambda ; y ) = -\log C(\lambda) - \sum_{k=1}^K y_k \log \lambda_k \iff p(y;\lambda) = C(\lambda) \cdot \prod_{k=1}^K \lambda_k^{y_k},
\end{align}
where $C(\lambda)$ is the normalizing constant: 
\begin{align}
C(\lambda) = \left( \int_{\Delta^K} \prod_{k=1}^K \lambda_k^{y_k} d y_k \right)^{-1}.
\end{align}
This distribution was found to possess a number of attractive theoretical and empirical properties \cite{gordon2020continuous}; we highlight the closed form expression of its normalizing constant:
\begin{align} \label{eq:norm_const}
C(\lambda) = \left( 
(-1)^{K+1} \sum_{k=1}^K \frac
 { \lambda_k }
 {  \prod_{i\ne k} \log{\frac{\lambda_{i}}{\lambda_{k}}} } \right)^{-1}
, 
\end{align}
which enables the use of automatic differentiation for optimizing models with the continuous-categorical log-likelihood (\ref{eq:cts_xe_loss}).
We also highlight that the continuous-categorical outperformed the Dirichlet distribution in regression models of compositional data, including neural network models \cite{gordon2020continuous}.

\section{Continuous-Categorical Label Smoothing} \label{sec:ls}

Label smoothing \cite{szegedy2016rethinking} has enjoyed rapid growth and widespread use as a means to reduce overfitting and improve the out-of-sample accuracy of neural network classifiers across a range of tasks including computer vision \cite{zoph2018learning, real2019regularized}, speech recognition \cite{chorowski2016towards}, and machine translation \cite{vaswani2017attention}.
The mechanism is simple: given a neural network classifier $f_\theta : x \to y$, we replace our one-hot labels $y \in \Omega$ with ``soft'' targets:
\begin{align} \label{eq:ls}
y^{\text{LS}} = (1 - \varepsilon) y + \varepsilon u,
\end{align}
where $u = (1/K, \dots, 1/K)^\top$ is a uniform vector and $\varepsilon > 0$ is a constant.
The network weights $\theta$ are then trained to minimize the cross-entropy loss between the network output $f_\theta(x)$ and the modified data $y^{\text{LS}}$.

Equation \ref{eq:ls} maps $y \in \Omega$ \ to \ $y^{\text{LS}} \in \Delta^K$, so that our targets are no longer categorical, but simplex-valued. 
Thus, even though they are not continuously distributed, it is natural to consider label-smoothed classification through the lens of compositional regression.
Our proposal is therefore to use a continuous-categorical log-likelihood in lieu of the cross-entropy loss, and we refer to this model as CC-LS.
Namely, we are interested in comparing the usual label smoothing loss:
\begin{align}
\min_\theta \mathcal L^{\text{LS}} (\theta) = - \sum_{(x, y)} \sum_{k} y_k^{\text{LS}} \cdot \log [f_\theta(x)]_k
,
\end{align}
against its continuous-categorical counterpart:
\begin{align}
\min_\theta \mathcal L^{\text{CC-LS}} (\theta) = - \sum_{(x, y)} \left\{ \log C(f_\theta(x) ) + \sum_{k} y_k^{\text{LS}} \cdot \log [f_\theta(x)]_k \right\}
.
\end{align}



We remark that, strictly speaking, in order to make our targets continuous over the simplex, we would also have to add continuous noise to the labels, for example by drawing $u$ uniformly at random on the simplex.
Such an approach produced little difference over using the fixed value $u = (1/K, \dots, 1/K)^\top$, neither in LS nor CC-LS, and we will omit the results for clarity.
However, given the wealth of existing methods that achieve improved generalization error by adding noise at different stages in the training procedure \cite{bishop1995training, srivastava2014dropout, shorten2019survey}, the idea of smoothing the labels with random noise may still hold potential, and we leave its further analysis for future work.


\subsection{Experiments}

Following the experimental setup of Muller et al \cite{muller2019does}, we train a CNN classifier on CIFAR-10, with and without label smoothing as well as our novel CC-LS model (see appendix \ref{app:ls} for the full details of our architecture).
This is an example in which label smoothing provided no significant gain over the un-smoothed baseline, likely because the CNN is already regularized using dropout \cite{srivastava2014dropout}, weight decay \cite{krogh1992simple}, and batch normalization \cite{ioffe2015batch}, which altogether are sufficient to provide a good model of the data, given the level of complexity of CIFAR-10. 
Likewise, we find that the CC-LS model also performs no better than the baseline in this setting (top row of Table \ref{tab:ls}).

Driven by these observations, we perform an ablation study over the different regularizers used in our network, and the results paint a more interesting picture (Table \ref{tab:ls}).
Notably, we find that for the unregularized CNN (bottom row), CC-LS significantly outperforms both LS and the baseline.
In the case where our network is partially regularized with dropout only, the baseline becomes equally good as LS, but the gap with CC-LS remains wide (penultimate row), and the gain from CC-LS persists after adding weight decay.
On the other hand, batch normalization (top half) was sufficient to capture all the gain in test accuracy, with neither LS nor CC-LS outperforming the baseline in these cases.
Under weight decay without batch normalization (rows 5 and 6), training became less stable (as evidenced by the large standard deviations), but CC-LS was able to reduce the variability in model accuracy.
Overall, Table \ref{tab:ls} indicates that the CC-LS loss function provides a different (and sometimes, significantly better) regularization effect than that of vanilla LS, suggesting its potential for novel applications, particularly in the settings where batch normalization may be undesirable \cite{galloway2019batch}.
We note further that numerous existing works have been devoted to analyzing the interplay between dropout, weight decay, and batch normalization \cite{van2017l2, garbin2020dropout, chen2019rethinking, li2019understanding, hernandez2018deep}; our focus is specifically on their relation to label smoothing and CC-LS.


%

\begin{table}
  \caption{Ablation study for label smoothing on CIFAR-10. We show out-of-sample accuracy for our baseline classifier (w/o LS), as well as vanilla LS and CC-LS, both with $\varepsilon=0.1$. Errors indicate the standard deviation over 10 random initializations of the network. We consider the effect of LS and CC-LS over the baseline under each combination of dropout, weight decay and batch normalization, and find that CC-LS provides significant outperformance in the absence of BatchNorm.} \label{tab:ls}
  \centering
  \begin{tabular}{cccccc}
    \toprule
    Dropout & Weight decay & BatchNorm & w/o LS & with LS & CC-LS \\
    \midrule
    Yes & Yes & Yes & $\boldsymbol{89.5} \ (\pm 0.1)$ & ${89.1} \ (\pm 0.2)$ & $89.0 \ (\pm 0.2)$ \\
    No & Yes & Yes & $\boldsymbol{89.6} \ (\pm 0.1)$ & ${89.2} \ (\pm 0.1)$ & $89.2 \ (\pm 0.2)$ \\
    Yes  & No & Yes & $\boldsymbol{89.4} \ (\pm 0.2)$ & $89.3 \ (\pm 0.2)$ & ${89.0} \ (\pm 0.2)$ \\
    No  & No &  Yes & $\boldsymbol{89.5} \ (\pm 0.2)$ & $89.4 \ (\pm 0.1)$ & ${89.1} \ (\pm 0.2)$ \\
    Yes & Yes & No & $88.6 \ (\pm 1.2)$ & ${88.6} \ (\pm 1.0)$ & ${88.7} \ (\pm 0.6)$ \\
    No & Yes & No & $88.8 \ (\pm 1.2)$ & ${88.7} \ (\pm 1.0)$ & ${88.6} \ (\pm 0.6)$ \\
    Yes  & No & No & $87.0 \ (\pm 0.2)$ & $87.0 \ (\pm 0.1)$ & $\boldsymbol{87.6} \ (\pm 0.2)$ \\
    No  & No &  No & $86.8 \ (\pm 0.1)$ & $87.0 \ (\pm 0.2)$ & $\boldsymbol{87.6} \ (\pm 0.2)$ \\
    \bottomrule
  \end{tabular}
\end{table}

\begin{wraptable}{r}{0.55\textwidth}
  \caption{Ratio of within-cluster sum of squares over between-cluster sum of squares, for the learned representations of Figure \ref{fig:ls}. Each cell shows the mean ratio over 10 random initializations, with standard errors.} \label{tab:ls_rep}
  \centering
\begin{tabular}{cccc}\\ \toprule  
Samples & w/o LS & with LS & CC-LS \\ \midrule
Training & $18\%\ (\pm 1)$ & $9\%\ (\pm 1)$ & $12\%\ (\pm 1)$\\  
Test & $25\%\ (\pm 1)$ & $20\%\ (\pm 1)$ & $23\%\ (\pm 1)$ \\  \bottomrule
\end{tabular}
\end{wraptable}

\begin{figure*} 
\centering
  \includegraphics[width=1.0\linewidth]{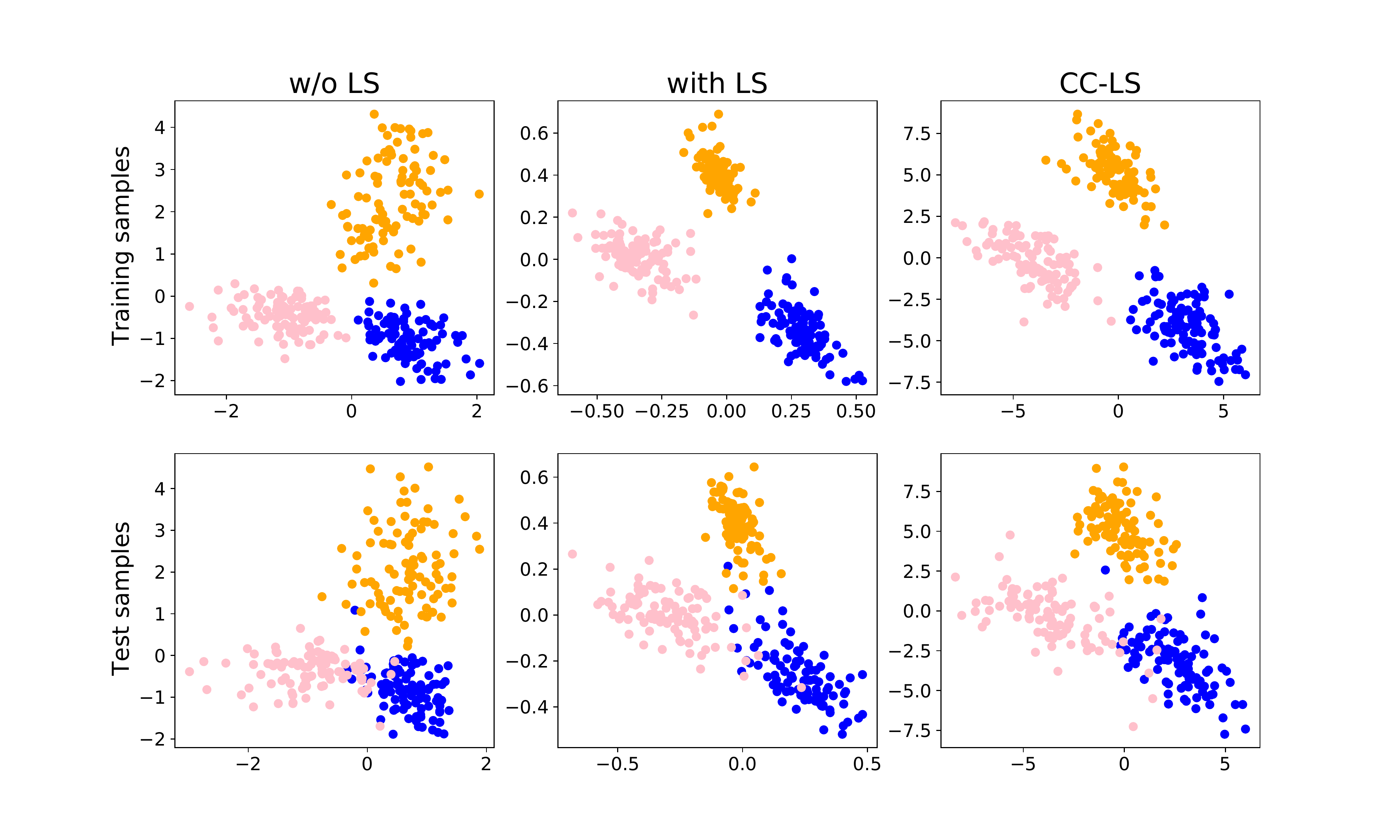}
  \caption{Learned representations for the classes ``airplane'' (blue), ``automobile'' (orange), and ``bird'' (pink), projected to an informative 2-dimensional affine subspace spanning the template-vectors of the 3 classes. We show the same plot for samples from the training (above) and test set (below), for the unsmoothed baseline (left), LS (middle), and CC-LS (right), trained with regularization (following the top row of Table \ref{tab:ls}). Note that CC-LS does not concentrate clusters as tightly as LS, suggesting the potential for richer learned representations.} \label{fig:ls}
\end{figure*}

We end this section with a qualitative analysis of the learned representations from our trained classifiers.
Again following \cite{muller2019does}, we define the ``template'' vector of the $k$th class, $w_k$, as the weight vector from the last CNN layer that is associated to the $k$th class, so that in other words:
\begin{align}
[f_\theta(x)]_k = \frac{e^{w_{k}^\top z}}{\sum_{k'} e^{w^\top_{k'} z}},
\end{align}
where $z$ is a vector containing the activations from the penultimate layer.
We then fix three classes, and construct an orthonormal basis (consisting of two vectors) for the plane containing their three template vectors.
For each of the classes, we pick a random sample of input data belonging to that class and project their penultimate layer activations onto this plane.
The results are shown in Figure \ref{fig:ls} for the (arbitrarily chosen) classes ``airplane'', ``automobile'', and ``bird''.
As was noted by \cite{muller2019does}, while label smoothing can help the classifier achieve better accuracy on the test set, it comes at the cost of a less informative learned representation, as can be seen from the more concentrated centroids in the second column relative to the first.
On the other hand, CC-LS achieves a somewhat richer representation than vanilla LS, as can be observed from the greater within-cluster variances in the third column, which we quantify in Table \ref{tab:ls_rep}.
This suggests that the CC may offer additional potential for combining LS with teacher models in the context of knowledge distillation, a setting in which the concentrated clusters enforced by LS proved detrimental to the training of a student model \cite{muller2019does}.

\section{Probabilistic Actor-Mimic Reinforcement Learning} \label{sec:rl}

In this section we summarize the Actor-Mimic Reinforcement Learning framework \cite{parisotto2015actor} and recast it as a compositional regression problem, highlighting the potential for a probabilistic model with the continuous-categorical distribution.

Actor-Mimic Networks (AMN) provide a method for multitask and transfer reinforcement learning. 
The goal of the AMN is to train a single agent to perform on several different ``source games'', $\{G_1, \dots, G_L\}$, each of which corresponds to a Markov Decision Process defined on a common state space and action set (shared across source tasks), but driven by a different set of transition probabilities and reward functions (specific to each task).
Formally, $G_i = (\mathcal S, \mathcal A, \mathcal T_i, \mathcal R_i)$, where $\mathcal S$ is the set of states, $\mathcal A$ is the set of actions, $\mathcal T_i(s'|s, a)$ is the probability of transitioning from state $s$ to state $s'$ when executing action $a$ in game $G_i$, and $\mathcal R_i$ is the reward function mapping states and actions to real-valued rewards representing the score of the $i$th game. 
In practice, each $G_i$ corresponds to a different videogame from the Atari Learning Environment \cite{bellemare2013arcade}; each game follows a different set of rules ($\mathcal T_i$ and $\mathcal R_i$) while taking place on the same console display ($\mathcal S$) and controller ($\mathcal A$).

In order to train an AMN, we first require access to a set of ``experts'' $\{ E_1, \dots, E_L \}$, each of which corresponds to an agent specialized in one of the source games.
Expert $E_i$ represents a policy $\pi_{E_i}$ mapping states to distributions over actions, so that we can write $\pi_{E_i}(a_k | s_t)$ for the probability that $E_i$ chooses action $a_k \in \mathcal A$ when in state $s_t \in \mathcal S$.
Note that $k$ indexes the action space, so that $\mathcal A = \{ a_1, \dots, a_K \}$, whereas $t$ indexes time (i.e., frame number), so that $\mathcal S \supseteq \{ s_1, s_2, \dots \}$.
In our implementation, $E_i$ corresponds to a Deep Q-Network (DQN) \cite{mnih2015human} trained on game $G_i$, though the fact that $E_i$ is a DQN is not necessary -- any policy that performs well on $G_i$ will suffice.


Given the set of expert policies, the AMN is trained to ``mimic'' the experts in their respective source games.
We can reformulate the method as a two-stage process. 
First, we form an auxiliary dataset of ``guidance vectors'', $\mathcal D_{\text{aux}} = \left\{ y_t^{(i)} \right\}$.
These vectors are obtained by generating, for each game $G_i$, a sequence of states $\left\{ s_t^{(i)} \right\}_{t=1}^n$, and then feeding these states through the corresponding expert policies, in other words: 
\begin{align}
y_t^{(i)} = \left(\pi_{E_i}\left(a_1|s_t^{(i)}\right), \dots, \pi_{E_i}\left(a_K | s_t^{(i)}\right) \right)
.
\end{align}
Second, the parameters of our Actor-Mimic Network, $\pi_\theta^{\text{AM}}$, are learned by minimizing the categorical cross-entropy loss with respect to the auxiliary data:
\begin{align} 
\min_\theta \mathcal L^{\text{AMN}}(\theta) 
&= - \sum_{t, i} \sum_{k} \pi_{E_i} \left(a_k|s_t^{(i)}\right) \cdot \log  \pi_\theta^{\text{AM}} \left( a_k | s_t^{(i)} \right). \label{eq:amn_obj}
\end{align}
In practice, we minimize this loss using minibatch stochastic gradient descent, running the gameplay-generation in parallel with the gradient steps.
The effectiveness of the AMN approach is fundamentally computational; the expert policies can be trained independently in parallel, and the AMN is much faster to optimize via the cross-entropy loss (\ref{eq:amn_obj}) than using policy gradients, as it is able to leverage the rich information from the expert policies directly, (with an entire guidance vector of probabilities containing information for all classes at each time step, rather than learning from noisy and biased $n$-step bootstrap estimates).

Since $y_t^{(i)}$ is a vector of probabilities over actions, our auxiliary data is simplex-valued rather than categorical.
It is therefore clear from Equation \ref{eq:amn_obj} that the AMN model solves a compositional regression problem, whence we propose replacing the cross-entropy loss with its probabilistic counterpart, the continuous-categorical log-likelihood:
\begin{align} \label{eq:CC-AMN_obj}
\min_\theta \mathcal L^{\text{CC-AMN}}(\theta) &= - \sum_{t, i} \left\{ \log C\left(\lambda_\theta^{\text{AM}}\left(s_t^{(i)}\right)\right) + \sum_{k} \pi_{E_i} \left(a_k|s_t^{(i)}\right) \cdot \log  \lambda_\theta^{\text{AM}}\left(a_k | s_t^{(i)}\right) \right\}.
\end{align}
We call this model CC-AMN, and we compare its performance against the AMN model, as well as the DQN baseline.

\subsection{Experiments} \label{sec:rl_exp}

We follow the experimental setup of Parisotto et al \cite{parisotto2015actor}, choosing a subset of games from the Atari Learning Environment in which the DQN model performed at super-human level.
For each game, we pre-train a DQN with the same network architecture; these are then used as the expert policies.
Our network architecture, described in appendix \ref{app:rl}, 
is taken directly from \cite{mnih2015human}, and is also used for the AMN and CC-AMN models.

First, we reproduce the results of \cite{parisotto2015actor} and compare with our novel CC-AMN model, as shown in Table \ref{tab:CC-AMN}.
The evaluation scores of the CC-AMN are similar to those of the AMN, except for the game of Pong, where using the CC likelihood leads to unstable training, resulting in worse performance and higher variability in the evaluation score.
Note that both AMN and CC-AMN are generally able to achieve similar performance to the expert DQN.

\begin{table}
  \caption{Mean evaluation score (and standard deviation) over the last 20 evaluation epochs (higher is better). With the exception of Pong, the performance of AMN and CC-AMN is similar.} \label{tab:CC-AMN}
  \label{sample-table}
  \centering
  \begin{tabular}{ccccc}
    \toprule
    Model & Breakout & Atlantis & Pong & SpaceInvaders \\
    \midrule
    DQN  & $331 \ (\pm 44)$ & $32\,833 \ (\pm 14\,430)$ & $20.9 \ (\pm 0.2)$ & $442 \ (\pm 119)$  \\
    AMN  & $337 \ (\pm 74)$ & $31\,558 \ (\pm 9\,084)$ & $20.9 \ (\pm 0.1)$ & $415 \ (\pm 126)$  \\
    CC-AMN  & $320 \ (\pm 66)$ & $26\,196 \ (\pm 10\,396)$ & $8.8 \ (\pm 11.9)$ & $415 \ (\pm 132)$ \\
    \bottomrule
  \end{tabular}
\end{table}

\begin{figure*} 
\centering
  \includegraphics[width=1.0\linewidth]{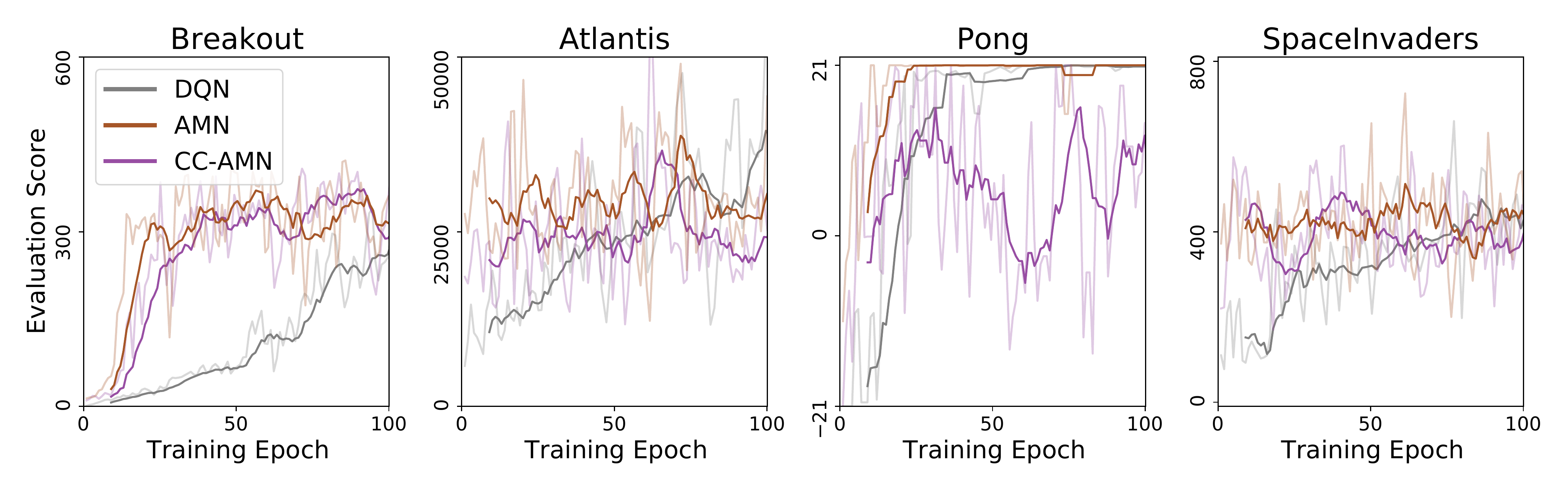}
  \caption{Training curves for the CC-AMN and AMN models, run on the simplified single-game objective. Solid lines reflect a moving average of the raw evaluation scores (faded lines). Each training epoch lasts $100\,000$ frames, with the evaluation scores being calculated from another $100\,000$ frames. While the actor-mimic models learn much faster than the DQN, the CC-AMN shows no improvement over the AMN model.} \label{fig:CC-AMN}
\end{figure*}

Second, we focus specifically on the effect of the probabilistic objective (\ref{eq:CC-AMN_obj}) on network training by reducing the multi-task objective to a single-task objective, i.e., we no longer sum over $i$ in Equation \ref{eq:amn_obj}.
This corresponds to running AMN and CC-AMN against the expert DQN, $E_i$, of a single game, and we do this separately for each game, as shown in Figure \ref{fig:CC-AMN}.
While both CC-AMN and AMN are able to train much faster than the DQN, converging in just a few epochs, CC-AMN fails to outperform AMN, and can be slower to converge (Breakout) or worse overall (Pong).

The case of Pong highlights an important failure mode of CC-AMN, which also offers some insight as to why our model underperforms in the other games.
The issue originates in the normalizing constant (\ref{eq:norm_const}), which is numerically unstable when the parameter $\lambda$ is close to uniform, due to the product of log-ratios vanishing in the in the denominator, as was noted in \cite{gordon2020continuous}.
In the case of CC-AMN, our optimization hovers around this unstable region, since the guidance vectors tend to concentrate around the centroid of the simplex (this is because our expert policies correspond to the softmax of Q-value functions, which don't typically exhibit large variability across actions since, over small time steps, most actions are not individually critical to the outcome of the game).
In practice this is not necessarily a problem, as we zero out the unstable gradients during optimization, as in \cite{gordon2020continuous}.
Doing so provides a reasonable approximation, since $\nabla G(\lambda) = 0$ for $\lambda = (1/K, \dots, 1/K)$, by symmetry ($\sum_k \lambda_k = 1$ is constrained by definition of the continuous-categorical).
Nevertheless, this behavior likely results in a worse optimization landscape overall (unlike in CC-LS where $y$, and hence $\lambda$, are far away from the centroid), which in the game of Pong, derails our gradient search altogether.
Further investigation may involve re-running our experiments with arbitrary-precision floating point, though we currently find this to be computationally prohibitive. 

%

\section{Discussion and Future Work} \label{sec:conclusion}

Comparing our experiments on CC-LS and CC-AMN, it may come as a surprise that the former yields the more promising empirical results, in spite of its less rigorous theoretical underpinning (with targets that are simplex-valued, but not genuinely continuous).
This observation suggests that the continuous-categorical may provide useful modeling advances outside the realm of compositional data analysis and probabilistic modeling, for example in classification problems.

It is also worth noting that, throughout our experiments, the cross-entropy loss may have benefitted disproportionately from favorable network initialization, which has been developed for and become increasingly specialized toward networks with particular loss functions \cite{he2015delving}.
A similar argument can be made about network architecture, noting that a good architecture for the cross-entropy loss may not be equivalent to a good architecture for its continuous-categorical counterpart.
In fact, we have observed experimentally that additional architecture or hyperparameter search can lead to improved performance for the CC-based approaches.
However we deliberately chose not to focus on such experimentation, as doing so could further entangle the effect of the loss function on our models; we leave such analyses to future work.

At a more theoretical level, as identified in \cite{gordon2020continuous},  it follows from the properties of exponential families that optimizing the continuous-categorical log-likelihood results in an unbiased estimator for its mean parameter, which corresponds to a (local) average of the observed data.
This is, in fact, akin to the cross-entropy loss, which is also maximized at local average that approximates the conditional expectation of the outputs given the inputs.
The optimization landscapes defined by the two loss functions could therefore be of similar nature, though this remains an open question.

Last, we highlight a computational limitation of our approach: the numerical instabilities noted in section \ref{sec:rl_exp} are exacerbated in high dimensions.
In fact, evaluating Equation \ref{eq:norm_const} for much more than 10 classes is problematic for all $\lambda \in \Delta^K$ (not only for $\lambda$ around the centroid), since the $K$ summands typically cancel out beyond numerical precision.\footnote{For an intuitive explanation, note that the Lebesgue measure of the $K$-dimensional simplex is $1/K!$. We therefore expect $C(\lambda)^{-1} \sim 1/K!$. However, as $K$ increases, the ``typical'' summand in (\ref{eq:norm_const}) decays slower than $1/K!$ (if at all). Thus, for large $K$, the individual summands in (\ref{eq:norm_const}) will become much larger (in magnitude) than $C(\lambda)^{-1}$. Their summation will then result in (near) total cancellation and therefore total loss of numerical precision.}
As a result, we constrained our experimentation to examples where $K \le 10$, however, similar applications with $K \sim 100$ or greater are also of interest.
Further advances in theory or numerical analysis will be needed to enable successful applications of the continuous-categorical at this scale.

We conclude by noting that, taken together with the theoretical and empirical results in \cite{loaiza2019continuous} and \cite{gordon2020continuous}, our work suggests that future methodological advances may be possible through a combination of careful probabilistic consideration of the cross-entropy loss, and the use of the continuous-categorical distribution.

\begin{ack}
We thank Andres Potapczynski and the anonymous reviewers for helpful conversations, and the
Simons Foundation, Sloan Foundation, McKnight Endowment Fund,
NSF 1707398, and the Gatsby Charitable Foundation for support.
\end{ack}

\bibliography{ref}

\begin{thebibliography}{31}
\providecommand{\natexlab}[1]{#1}
\providecommand{\url}[1]{\texttt{#1}}
\expandafter\ifx\csname urlstyle\endcsname\relax
  \providecommand{\doi}[1]{doi: #1}\else
  \providecommand{\doi}{doi: \begingroup \urlstyle{rm}\Url}\fi

\bibitem[Aitchison(1982)]{aitchison1982statistical}
John Aitchison.
\newblock The statistical analysis of compositional data.
\newblock \emph{Journal of the Royal Statistical Society: Series B
  (Methodological)}, 44\penalty0 (2):\penalty0 139--160, 1982.

\bibitem[Aitchison(1994)]{aitchison1994principles}
John Aitchison.
\newblock Principles of compositional data analysis.
\newblock \emph{Lecture Notes-Monograph Series}, 24:\penalty0 73--81, 1994.
\newblock ISSN 07492170.

\bibitem[Aitchison(1999)]{aitchison1999logratios}
John Aitchison.
\newblock Logratios and natural laws in compositional data analysis.
\newblock \emph{Mathematical Geology}, 31\penalty0 (5):\penalty0 563--580,
  1999.

\bibitem[Bellemare et~al.(2013)Bellemare, Naddaf, Veness, and
  Bowling]{bellemare2013arcade}
Marc~G Bellemare, Yavar Naddaf, Joel Veness, and Michael Bowling.
\newblock The arcade learning environment: An evaluation platform for general
  agents.
\newblock \emph{Journal of Artificial Intelligence Research}, 47:\penalty0
  253--279, 2013.

\bibitem[Bishop(1995)]{bishop1995training}
Chris~M Bishop.
\newblock Training with noise is equivalent to tikhonov regularization.
\newblock \emph{Neural computation}, 7\penalty0 (1):\penalty0 108--116, 1995.

\bibitem[Campbell and Mosimann(1987)]{campbell1987multivariate}
G~Campbell and J~Mosimann.
\newblock Multivariate methods for proportional shape.
\newblock In \emph{ASA Proceedings of the Section on Statistical Graphics},
  volume~1, pages 10--17. Washington, 1987.

\bibitem[Chen et~al.(2019)Chen, Chen, Shi, Hsieh, Liao, and
  Zhang]{chen2019rethinking}
Guangyong Chen, Pengfei Chen, Yujun Shi, Chang-Yu Hsieh, Benben Liao, and
  Shengyu Zhang.
\newblock Rethinking the usage of batch normalization and dropout in the
  training of deep neural networks.
\newblock \emph{arXiv preprint arXiv:1905.05928}, 2019.

\bibitem[Chorowski and Jaitly(2016)]{chorowski2016towards}
Jan Chorowski and Navdeep Jaitly.
\newblock Towards better decoding and language model integration in sequence to
  sequence models.
\newblock \emph{arXiv preprint arXiv:1612.02695}, 2016.

\bibitem[Egozcue et~al.(2003)Egozcue, Pawlowsky-Glahn, Mateu-Figueras, and
  Barcelo-Vidal]{egozcue2003isometric}
Juan~Jos{\'e} Egozcue, Vera Pawlowsky-Glahn, Gl{\`o}ria Mateu-Figueras, and
  Carles Barcelo-Vidal.
\newblock Isometric logratio transformations for compositional data analysis.
\newblock \emph{Mathematical Geology}, 35\penalty0 (3):\penalty0 279--300,
  2003.

\bibitem[Galloway et~al.(2019)Galloway, Golubeva, Tanay, Moussa, and
  Taylor]{galloway2019batch}
Angus Galloway, Anna Golubeva, Thomas Tanay, Medhat Moussa, and Graham~W
  Taylor.
\newblock Batch normalization is a cause of adversarial vulnerability.
\newblock \emph{arXiv preprint arXiv:1905.02161}, 2019.

\bibitem[Garbin et~al.(2020)Garbin, Zhu, and Marques]{garbin2020dropout}
Christian Garbin, Xingquan Zhu, and Oge Marques.
\newblock Dropout vs. batch normalization: an empirical study of their impact
  to deep learning.
\newblock \emph{Multimedia Tools and Applications}, pages 1--39, 2020.

\bibitem[Gordon-Rodriguez et~al.(2020)Gordon-Rodriguez, Loaiza-Ganem, and
  Cunningham]{gordon2020continuous}
Elliott Gordon-Rodriguez, Gabriel Loaiza-Ganem, and John~P Cunningham.
\newblock The continuous categorical: a novel simplex-valued exponential
  family.
\newblock In \emph{International Conference on Machine Learning}, 2020.

\bibitem[He et~al.(2015)He, Zhang, Ren, and Sun]{he2015delving}
Kaiming He, Xiangyu Zhang, Shaoqing Ren, and Jian Sun.
\newblock Delving deep into rectifiers: Surpassing human-level performance on
  imagenet classification.
\newblock In \emph{Proceedings of the IEEE international conference on computer
  vision}, pages 1026--1034, 2015.

\bibitem[Hern{\'a}ndez-Garc{\'\i}a and K{\"o}nig(2018)]{hernandez2018deep}
Alex Hern{\'a}ndez-Garc{\'\i}a and Peter K{\"o}nig.
\newblock Do deep nets really need weight decay and dropout?
\newblock \emph{arXiv preprint arXiv:1802.07042}, 2018.

\bibitem[Hijazi and Jernigan(2009)]{hijazi2009modelling}
Rafiq~H Hijazi and Robert~W Jernigan.
\newblock Modelling compositional data using dirichlet regression models.
\newblock \emph{Journal of Applied Probability \& Statistics}, 4\penalty0
  (1):\penalty0 77--91, 2009.

\bibitem[Hinton et~al.(2015)Hinton, Vinyals, and Dean]{hinton2015distilling}
Geoffrey Hinton, Oriol Vinyals, and Jeffrey Dean.
\newblock Distilling the knowledge in a neural network.
\newblock In \emph{NIPS Deep Learning and Representation Learning Workshop},
  2015.

\bibitem[Ioffe and Szegedy(2015)]{ioffe2015batch}
Sergey Ioffe and Christian Szegedy.
\newblock Batch normalization: Accelerating deep network training by reducing
  internal covariate shift.
\newblock \emph{arXiv preprint arXiv:1502.03167}, 2015.

\bibitem[Krogh and Hertz(1992)]{krogh1992simple}
Anders Krogh and John~A Hertz.
\newblock A simple weight decay can improve generalization.
\newblock In \emph{Advances in neural information processing systems}, pages
  950--957, 1992.

\bibitem[Li et~al.(2019)Li, Chen, Hu, and Yang]{li2019understanding}
Xiang Li, Shuo Chen, Xiaolin Hu, and Jian Yang.
\newblock Understanding the disharmony between dropout and batch normalization
  by variance shift.
\newblock In \emph{Proceedings of the IEEE conference on computer vision and
  pattern recognition}, pages 2682--2690, 2019.

\bibitem[Loaiza-Ganem and Cunningham(2019)]{loaiza2019continuous}
Gabriel Loaiza-Ganem and John~P Cunningham.
\newblock The continuous bernoulli: fixing a pervasive error in variational
  autoencoders.
\newblock In \emph{Advances in Neural Information Processing Systems}, pages
  13266--13276, 2019.

\bibitem[Mnih et~al.(2015)Mnih, Kavukcuoglu, Silver, Rusu, Veness, Bellemare,
  Graves, Riedmiller, Fidjeland, Ostrovski, et~al.]{mnih2015human}
Volodymyr Mnih, Koray Kavukcuoglu, David Silver, Andrei~A Rusu, Joel Veness,
  Marc~G Bellemare, Alex Graves, Martin Riedmiller, Andreas~K Fidjeland, Georg
  Ostrovski, et~al.
\newblock Human-level control through deep reinforcement learning.
\newblock \emph{nature}, 518\penalty0 (7540):\penalty0 529--533, 2015.

\bibitem[M{\"u}ller et~al.(2019)M{\"u}ller, Kornblith, and
  Hinton]{muller2019does}
Rafael M{\"u}ller, Simon Kornblith, and Geoffrey~E Hinton.
\newblock When does label smoothing help?
\newblock In \emph{Advances in Neural Information Processing Systems}, pages
  4694--4703, 2019.

\bibitem[Parisotto et~al.(2015)Parisotto, Ba, and
  Salakhutdinov]{parisotto2015actor}
Emilio Parisotto, Jimmy~Lei Ba, and Ruslan Salakhutdinov.
\newblock Actor-mimic: Deep multitask and transfer reinforcement learning.
\newblock \emph{arXiv preprint arXiv:1511.06342}, 2015.

\bibitem[Real et~al.(2019)Real, Aggarwal, Huang, and Le]{real2019regularized}
Esteban Real, Alok Aggarwal, Yanping Huang, and Quoc~V Le.
\newblock Regularized evolution for image classifier architecture search.
\newblock In \emph{Proceedings of the aaai conference on artificial
  intelligence}, volume~33, pages 4780--4789, 2019.

\bibitem[Shorten and Khoshgoftaar(2019)]{shorten2019survey}
Connor Shorten and Taghi~M Khoshgoftaar.
\newblock A survey on image data augmentation for deep learning.
\newblock \emph{Journal of Big Data}, 6\penalty0 (1):\penalty0 60, 2019.

\bibitem[Srivastava et~al.(2014)Srivastava, Hinton, Krizhevsky, Sutskever, and
  Salakhutdinov]{srivastava2014dropout}
Nitish Srivastava, Geoffrey Hinton, Alex Krizhevsky, Ilya Sutskever, and Ruslan
  Salakhutdinov.
\newblock Dropout: a simple way to prevent neural networks from overfitting.
\newblock \emph{The journal of machine learning research}, 15\penalty0
  (1):\penalty0 1929--1958, 2014.

\bibitem[Szegedy et~al.(2016)Szegedy, Vanhoucke, Ioffe, Shlens, and
  Wojna]{szegedy2016rethinking}
Christian Szegedy, Vincent Vanhoucke, Sergey Ioffe, Jon Shlens, and Zbigniew
  Wojna.
\newblock Rethinking the inception architecture for computer vision.
\newblock In \emph{Proceedings of the IEEE conference on computer vision and
  pattern recognition}, pages 2818--2826, 2016.

\bibitem[Tzeng et~al.(2015)Tzeng, Hoffman, Darrell, and
  Saenko]{tzeng2015simultaneous}
Eric Tzeng, Judy Hoffman, Trevor Darrell, and Kate Saenko.
\newblock Simultaneous deep transfer across domains and tasks.
\newblock In \emph{Proceedings of the IEEE International Conference on Computer
  Vision}, pages 4068--4076, 2015.

\bibitem[Van~Laarhoven(2017)]{van2017l2}
Twan Van~Laarhoven.
\newblock L2 regularization versus batch and weight normalization.
\newblock \emph{arXiv preprint arXiv:1706.05350}, 2017.

\bibitem[Vaswani et~al.(2017)Vaswani, Shazeer, Parmar, Uszkoreit, Jones, Gomez,
  Kaiser, and Polosukhin]{vaswani2017attention}
Ashish Vaswani, Noam Shazeer, Niki Parmar, Jakob Uszkoreit, Llion Jones,
  Aidan~N Gomez, {\L}ukasz Kaiser, and Illia Polosukhin.
\newblock Attention is all you need.
\newblock In \emph{Advances in neural information processing systems}, pages
  5998--6008, 2017.

\bibitem[Zoph et~al.(2018)Zoph, Vasudevan, Shlens, and Le]{zoph2018learning}
Barret Zoph, Vijay Vasudevan, Jonathon Shlens, and Quoc~V Le.
\newblock Learning transferable architectures for scalable image recognition.
\newblock In \emph{Proceedings of the IEEE conference on computer vision and
  pattern recognition}, pages 8697--8710, 2018.

\end{thebibliography}
\bibliographystyle{plain}

\newpage
\appendix

\section{Experimental details}
\subsection{Label Smoothing} \label{app:ls}

We denote a convolutional layer by W$\times$W$\times$N-S, where W is the width of the convolution, N the number of filter maps, and S the stride. 
Our architecture is: 3$\times$3$\times$32-1 $\to$ BatchNorm $\to$ 3$\times$3$\times$32-1 $\to$ BatchNorm $\to$ MaxPooling (2$\times$2) $\to$ Dropout (0.2) $\to$ 3$\times$3$\times$64-1 $\to$ BatchNorm $\to$ 3$\times$3$\times$64-1 $\to$ BatchNorm $\to$ MaxPooling (2$\times$2) $\to$ Dropout (0.3) $\to$ 3$\times$3$\times$128-1 $\to$ BatchNorm $\to$ 3$\times$3$\times$128-1 $\to$ BatchNorm $\to$ MaxPooling (2$\times$2) $\to$ Dropout (0.4) $\to$ 10 fully-connected units.
We use weight decay of 0.0001 in the final fully-connected layer, which, together with dropout and batch normalization, was switched on and off in the different runs of our ablation study in Table \ref{tab:ls}.
Our models were trained for 500 epochs using a minibatch size of 128 and the Adam optimizer with a learning rate of $10^{-3}$.
The label smoothing hyperparameter $\varepsilon$ was set to 0.1 as per \cite{muller2019does}.

Note however, that we were unable to replicate the results of \cite{muller2019does} exactly, as they did not share their code, nor did they describe their architecture in full.

\subsection{Actor-Mimic Network} \label{app:rl}

Our architecture is 8$\times$8$\times$32-4 $\to$ 4$\times$4$\times$64-2 $\to$ 3$\times$3$\times$64-1 $\to$  7$\times$7$\times$1024-1 $\to$ 512 fully-connected units $\to$ 6 fully connected units (corresponding to 6 possible actions).
We used the Adam optimizer with a learning rate of $10^{-5}$, and a minibatch size of 32.

\end{document}